\documentclass[11pt]{article}

\usepackage[preprint]{acl}

\usepackage{times}
\usepackage{latexsym}

\usepackage[T1]{fontenc}
\usepackage[utf8]{inputenc}

\usepackage{microtype}

\usepackage{inconsolata}

\usepackage{graphicx}

\usepackage{enumitem}
\usepackage{graphicx}
\usepackage[utf8]{inputenc} % allow utf-8 input
\usepackage[T1]{fontenc}    % use 8-bit T1 fonts
\usepackage{hyperref}       % hyperlinks
\usepackage{url}            % simple URL typesetting
\usepackage{booktabs}       % professional-quality tables
\usepackage{amsfonts}       % blackboard math symbols
\usepackage{nicefrac}       % compact symbols for 1/2, etc.
\usepackage{microtype}      % microtypography
\usepackage{xcolor}         % colors
\usepackage[table]{xcolor}
\usepackage{array}
\usepackage{amssymb}
\usepackage{enumitem}
\usepackage{graphicx}
\usepackage{subcaption}
\usepackage{threeparttable}
\usepackage{pifont}
\usepackage{amsmath}
\usepackage{multirow}

\usepackage{listings}
\usepackage{makecell}
\usepackage{hyperref}
\usepackage{url}
\usepackage{fontawesome5}

\definecolor{red}{RGB}{238, 68, 51}
\definecolor{blue}{RGB}{70, 177, 225}
\definecolor{yellow}{RGB}{255, 192, 0}
\definecolor{purple}{RGB}{216, 110, 204}
\definecolor{brown}{RGB}{127, 36, 28}
\definecolor{green}{RGB}{71, 172, 20}
\definecolor{orange}{RGB}{194,153,107}

\newcolumntype{C}[1]{>{\centering\let\newline\\\arraybackslash\hspace{0pt}}m{#1}}

\newcolumntype{L}[1]{>{\raggedright\arraybackslash}m{#1}}

\title{UI-KOBE: Knowledge-Oriented Behavior Exploration for Lightweight Graph-Guided GUI Agents}

\author{
 \textbf{Yuxiang Chai\textsuperscript{1}},
 \textbf{Han Xiao\textsuperscript{1}},
 \textbf{Xinyu Fu\textsuperscript{2}},
 \textbf{Jinpeng Chen\textsuperscript{2}},
 \textbf{Rui Liu\textsuperscript{2}},
 \textbf{Hongsheng Li\textsuperscript{1,3,4 \textdagger}}
\\
 \small
 \textsuperscript{1}CUHK MMLab,
 \textsuperscript{2}Huawei Research,
 \textsuperscript{3}Shenzhen Loop Area Institute,
 \textsuperscript{4}CPII under InnoHK,
 \textsuperscript{\textdagger}Corresponding author
 \\
 \small
 \faGithub \quad \url{https://github.com/YuxiangChai/UI-KOBE}
}

\begin{document}
\maketitle

\begin{abstract}
Recent advances in mobile GUI agents have shown strong potential for automating mobile tasks, but most effective systems still depend on large vision-language models for screenshot understanding and long-horizon planning. Small GUI agents that can be deployed directly on mobile devices are more attractive for practical use, offering lower inference cost and better protection of sensitive on-device information. However, due to limited model capacity, such lightweight agents remain unreliable when planning and executing GUI tasks end-to-end from screenshots alone. We propose Knowledge-Oriented Behavior Exploration (\textbf{UI-KOBE}), a framework that improves lightweight mobile GUI agents with reusable app-specific graph knowledge. UI-KOBE first autonomously explores a mobile application and constructs an app knowledge graph, where nodes represent distinct UI states and edges represent executable transitions. At runtime, a lightweight GUI agent uses the graph as external guidance: given a user task and the current screenshot, it identifies the current graph node and selects among self-loop actions, neighboring transitions, task completion, or fallback free actions associated with that node. By supporting runtime decisions with app-specific graph guidance, UI-KOBE reduces the burden of end-to-end GUI planning and helps lightweight models perform mobile GUI tasks more effectively, offering a practical step toward efficient, interpretable, and privacy-conscious on-device GUI agents.
\end{abstract}

\section{Introduction}

Graphical User Interface (GUI) agents have recently shown strong potential for automating mobile and desktop tasks, driven by advances in vision-language models (VLMs) that can interpret screenshots and generate actions. Typically GUI interaction is an end-to-end formulation: given a task and the current screen, the model directly plans and executes a sequence of actions. While effective with large-scale proprietary or open-source models, this paradigm introduces two practical challenges. First, large open-source models require substantial computational resources, making deployment on-device difficult, while proprietary models requires high API costs. Second, smaller models that are suitable for on-device deployment, such as 4B-scale models, often struggle with long-horizon reasoning and planning, leading to unreliable task execution. Despite the limitation, lightweight GUI agents are highly desirable. They offer lower inference cost and better alignment with real-world deployment scenarios where sensitive user data can remain local. However, enabling small models to perform complex GUI tasks remains an open challenge. In particular, asking a small model to reason over the entire task at each step places a heavy burden on its limited capacity.

\begin{figure*}[ht]
    \centering
    \includegraphics[width=0.9\linewidth]{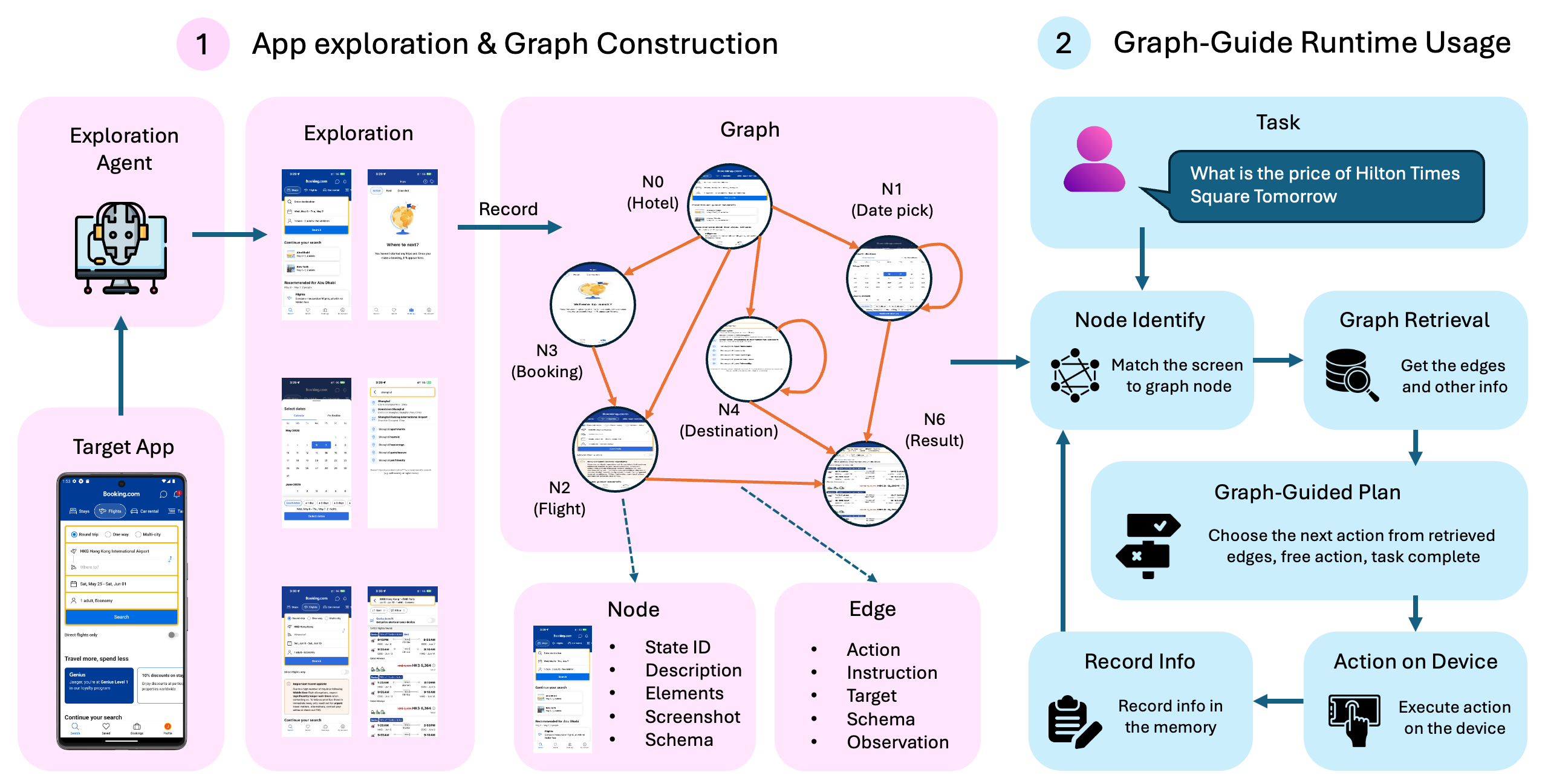}
    \caption{Overview of our framework. UI-KOBE first explores a target app and constructs an app knowledge graph, where nodes represent UI states and edges represent executable transitions. At runtime, a lightweight GUI agent uses the graph to identify the current node, retrieve local transition knowledge, select the next action, and execute the task step by step.}
    \label{fig:overview}
\end{figure*}

In this work, we argue that mobile GUI task execution should not rely solely on end-to-end reasoning at runtime. Instead, we propose to decouple \emph{app knowledge acquisition} from \emph{task-time execution}. We introduce Knowledge-Oriented Behavior Exploration (\textbf{UI-KOBE}), a framework that builds a reusable knowledge graph of an application through autonomous exploration, and this graph can be used to guide a runtime agent during task execution. Figure~\ref{fig:overview} illustrates the overall pipeline, including app exploration, graph construction, and graph-guided runtime usage. UI-KOBE represents an application as a directed graph, where nodes correspond to distinct UI states and edges correspond to transitions between states. The graph is constructed by an exploration agent that iteratively observes screens, executes actions, and records transitions. Each node captures semantic and structural information about a UI state, while each edge encodes both low-level actions and higher-level interaction patterns. Importantly, this graph represents general app behavior rather than specific task data, making it reusable across different tasks and users.

At runtime, the graph serves as an external knowledge source for a lightweight GUI agent. Instead of performing open-ended planning, the agent first identifies the current UI state within the graph and then selects the next action from a constrained set of graph-supported options, including self-loop operations and transitions to neighboring states. This formulation reduces GUI task execution to a sequence of guided local decisions, significantly lowering the reasoning burden on small models. When graph guidance is unavailable, the system falls back to a naive planner, ensuring robustness without reverting to full end-to-end reasoning. By leveraging pre-built app knowledge, UI-KOBE enables small models to perform GUI tasks more reliably. It also improves interpretability by grounding each action in explicit graph structures and supports reuse across tasks without repeated exploration.

This paper makes the following contributions:
\begin{itemize}[leftmargin=5mm]
    \item We propose a paradigm that decouples app knowledge acquisition from task-time execution, enabling graph-guided mobile GUI agents for lightweight models.
    \item We introduce UI-KOBE, a method for constructing a reusable app knowledge graph through autonomous exploration, including principled definitions of UI states (nodes) and executable transitions (edges).
    \item We design a graph-guided runtime agent that leverages local graph context to replace end-to-end planning with guided decision making, substantially improving the capability and reliability of small GUI agents.
\end{itemize}

\section{Related Work}

\subsection{End-to-End GUI Agents}
GUI agents aim to complete user tasks by perceiving graphical interfaces, reasoning over instructions, and executing actions such as clicking, typing, and swiping; recent surveys and benchmarks provide comprehensive overviews and evaluation resources for this rapidly growing area~\citep{wang2025guiagentsfoundationmodels,liu2025llmpoweredguiagentsphone,hu-etal-2025-os,chai-etal-2025-amex,chai2026a3androidagentarena,rawles2025androidworlddynamicbenchmarkingenvironment}. Most recent systems formulate GUI control as an end-to-end problem, where the model predicts actions directly from screenshots or UI representations and task instructions. Representative examples include UI-TARS~\cite{qin2025uitarspioneeringautomatedgui}, Mobile-Agent-V3/V3.5~\cite{ye2025mobileagentv3fundamentalagentsgui,xu2026mobileagentv35multiplatformfundamentalgui}, UI-Genie~\cite{xiao2025uigenieselfimprovingapproachiteratively}, UI-Venus~\cite{venusteam2026uivenus15technicalreport}, and MAI-UI~\cite{zhou2025maiuitechnicalreportrealworld}, which improve GUI grounding, planning, and execution through stronger foundation models, trajectory data, reinforcement learning, and model merging. Several works further study smaller GUI models, such as InfiGUI-R1~\cite{liu2025infiguir1advancingmultimodalgui}, UI-R1~\cite{lu2025uir1enhancingefficientaction}, Ferret-UI-Lite~\cite{yang2025ferretuilitelessonsbuilding}, and small variants of MAI-UI/UI-Venus, showing the promise of lightweight agents for efficient deployment. Different from these end-to-end approaches, our work reduces the runtime reasoning burden of lightweight agents by first constructing reusable app-specific graph knowledge and then using it to guide step-by-step decisions.

\subsection{Exploration-Based GUI Agents}

Recent work has also explored using app-specific knowledge, memory, or trajectory history to improve GUI task execution. AppAgent~\cite{zhang2023appagentmultimodalagentssmartphone} builds an app-level knowledge base from autonomous exploration and demonstrations, enabling agents to reuse prior interaction experience. AutoDroid~\cite{wen2024autodroidllmpoweredtaskautomation} constructs UI transition graphs (UTGs) through app exploration and uses them as structured app memory for mobile task automation. UI-Mem~\cite{xiao2026ui} introduces a memory mechanism that stores and reuses historical GUI interaction experience to improve long-horizon task execution and reduce repeated errors. KG-RAG~\cite{guan2025kgragenhancingguiagent} transforms fragmented UTGs into a vectorized knowledge database of intent-trajectory pairs, allowing agents to retrieve relevant navigation paths during online execution. GraphPilot~\cite{yu2026graphpilot} constructs app-specific knowledge graphs of page functions, element functions, and transition rules, and uses them to generate nearly complete action sequences with fewer LLM queries. Different from these methods, UI-KOBE focuses on building a semantic state-transition graph as a reusable behavioral abstraction, and uses it as a local decision scaffold for lightweight GUI agents: instead of retrieving an entire trajectory or generating a full action sequence, the runtime agent identifies the current node and selects the next graph-supported action step by step.

\section{UI-KOBE: Knowledge-Oriented Behavior Exploration}

\begin{figure*}[t]
    \centering
    \includegraphics[width=0.9\textwidth]{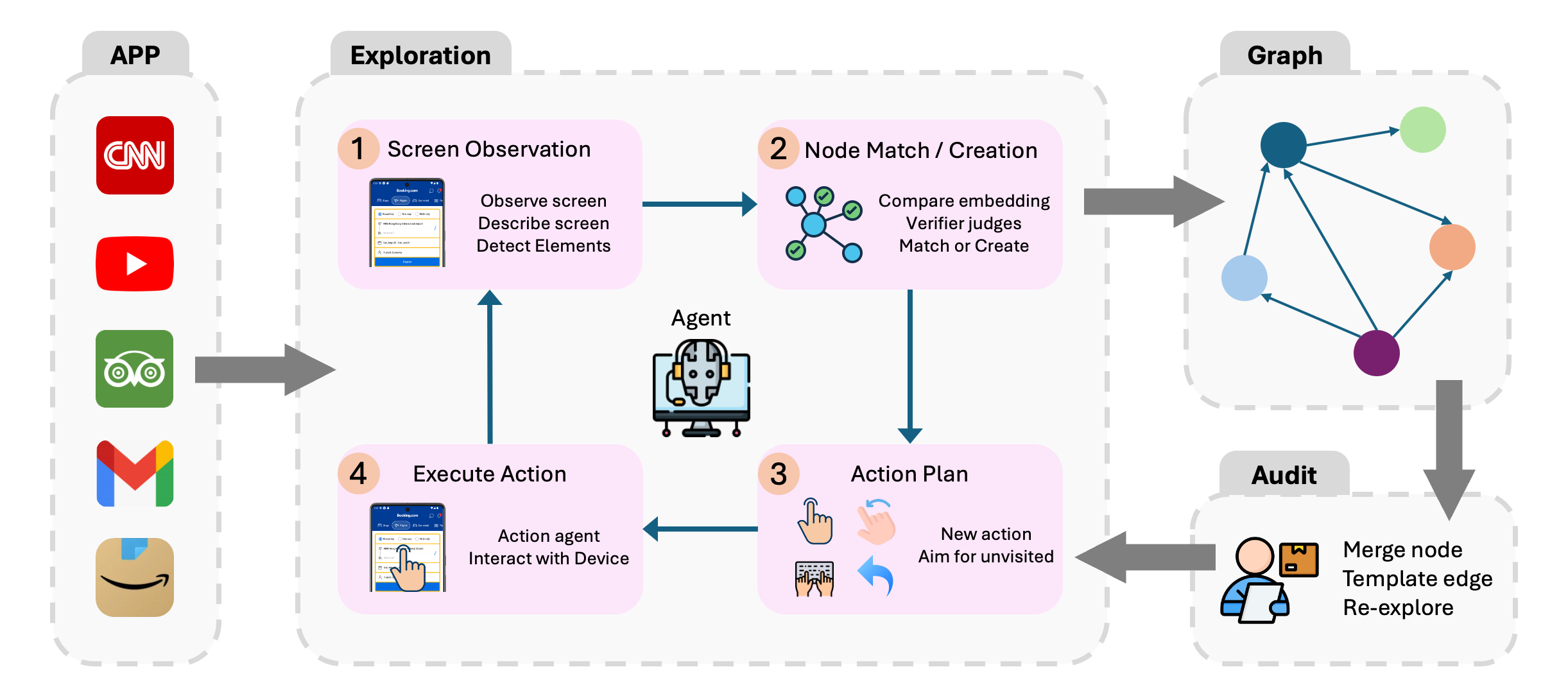}
    \caption{
    Overview of UI-KOBE. Given a target mobile app, the exploration agent repeatedly observes the current screen, matches or creates a graph node, plans an unexplored action, and executes it on the device. The observed transition is recorded into an app knowledge graph, which is further refined through auditing operations such as node merging, edge templating, and re-exploration.
    }
    \label{fig:uikobe_overview}
\end{figure*}

UI-KOBE is an app exploration method for constructing a reusable knowledge graph of a mobile application. Given a target app, UI-KOBE autonomously interacts with its interface, discovers UI states, records executable transitions, and incrementally builds a graph that captures app-level navigation and interaction knowledge. Figure~\ref{fig:uikobe_overview} illustrates the overall UI-KOBE pipeline, including screen observation, node matching or creation, action planning, action execution, graph construction, and post-hoc auditing. The resulting graph is not a task execution policy itself; rather, it serves as a reusable app-specific knowledge artifact that can later be used by a graph-guided GUI agent (Section~\ref{sec:graph_guided_agent}). This section focuses on how UI-KOBE defines, constructs, and refines the knowledge graph.

\subsection{Graph Representation}

Given a mobile application $\mathcal{A}$, UI-KOBE constructs a directed graph
\[
G_{\mathcal{A}} = (V, E),
\]
where each node $v \in V$ represents a semantic UI state and each edge $e \in E$ represents an observed executable transition between UI states. 

\paragraph{Node Definition.}
A node represents a distinct semantic UI state rather than an individual screenshot. Specifically, UI-KOBE abstracts a screen according to its functional role in the app, such as a search page, settings page, or search result page, while allowing dynamic screen contents to vary across visits. For example, search result pages produced by different queries may still correspond to the same node if they share the same function and layout. Conversely, visually similar screens with different roles, such as route departure selection and route destination selection pages, should be represented as different nodes. To support this abstraction, each node is associated with a semantic page description and auxiliary state information, such as visible dynamic values, a reference screenshot, and interactable elements. In this way, UI-KOBE treats node construction as a semantic state abstraction problem rather than simple screenshot matching.

\paragraph{Edge Definition.}
An edge represents an observed UI transition caused by an GUI interaction. Each edge stores the source node, target node, executed action json, natural-language instruction, and target observation. The target observation describes the effect of the action, such as navigating to another neighbor node or modifying the current screen state. Edges can connect different nodes, e.g., moving from a search page to a search result page, or form self-loops when the screen template remains unchanged. For self-loops, UI-KOBE records a schema delta that specifies which state variable or UI element changes, such as updating a query field or toggling a setting. Thus, edges encode both cross-screen navigation and within-screen state-transforming operations.

\subsection{Autonomous Exploration}

UI-KOBE constructs the graph through an iterative observe-identify-plan-act loop. At each exploration step, the agent observes the current screen, identifies the corresponding graph node, selects an unexplored interaction, executes one grounded device action, and enters another loop step. During observation and identification, the transition is also recorded into the graph.

\paragraph{Observation \& Identification.}
When a screenshot is observed, UI-KOBE first generates a semantic page description, a structured state snapshot, and the set of interactable elements. To identify whether the current screen corresponds to an existing node, UI-KOBE compares the embedding of the generated page description with stored embeddings of existing nodes in the same application. If the most similar candidate exceeds a threshold, UI-KOBE performs screenshot-level verification between the current screenshot and the candidate node's reference screenshot. This verification step prevents accidental merging of screens whose textual descriptions are similar but whose UI semantics differ. If the candidate is verified, the existing node is updated with the new observation; otherwise, UI-KOBE creates a new node with a fresh identifier, description, state snapshot, reference screenshot, and interactable elements.

\paragraph{Action Planning and Execution.}
After identifying the current node, UI-KOBE retrieves the outgoing edges that have already been explored and the visible elements that remain unexplored. A planner then proposes a natural-language instruction for the next interaction based on the current page description, existing outgoing transitions, and unexplored elements. The instruction is grounded into a single executable device action, such as tapping, typing, swiping, waiting, or pressing a system button. UI-KOBE then executes only one action per exploration step, making each recorded transition easier to interpret and failures easier to localize.

\paragraph{Transition Recording.}
After execution, UI-KOBE enters next step and observes the next screen and identifies its graph node using the same state identification procedure. It then records an edge from the previous node to the new node, including the executed action, planner instruction, target observation, and optional schema delta. If the source and target nodes are the same, the transition is treated as a self-loop and its state-changing effect is summarized through the schema delta. The graph is saved after each step, so exploration can resume from partial progress after interruptions.

\subsection{Graph Refinement and Re-Exploration}

The raw graph produced by autonomous exploration may contain duplicate nodes, wrong transitions, or uneven coverage. UI-KOBE therefore includes several refinement mechanisms to improve graph quality.

\paragraph{Graph Auditing.}
Autonomous exploration can produce noisy graph structures, such as duplicate nodes, incorrect merges, or abnormal transitions caused by mistaken actions and external-app jumps. UI-KOBE therefore performs a post-hoc audit over the raw graph. It detects suspicious node pairs using semantic similarity, reference screenshots, and overlapping outgoing actions, and verifies whether they represent the same UI state. Confirmed duplicates are merged, while functionally different screens are kept separate. The audit also flags unreliable edges whose target observations are inconsistent with the executed action or transition for later re-exploration.

\paragraph{Edge Normalization.}
Exploration naturally produces concrete instructions, such as typing a specific keyword or selecting a specific result. UI-KOBE normalizes similar instructions into reusable templates when possible. For instance, a concrete instruction like ``Type Starbucks'' can be abstracted into a parameterized instruction template for entering a query. This allows the graph to encode reusable interaction patterns rather than only one-off exploration traces.

\paragraph{Coverage-Oriented Re-Exploration.}
To avoid over-expanding only the most recent trajectory, UI-KOBE periodically selects under-explored nodes for continued exploration. The system can replay known transitions from a start node to reach a selected under-explored node and then continue exploring from that point. This coverage-oriented re-exploration improves the completeness of the graph and helps discover interactions that may be missed in a single linear exploration trajectory.

\section{Graph-Guided GUI Agent}
\label{sec:graph_guided_agent}

After UI-KOBE constructs an app knowledge graph, we use it to guide a runtime GUI agent during task execution. The motivation is to replace end-to-end GUI planning from screenshots with graph-guided decision making. At each step, the agent observes the current screen, identifies the corresponding graph node, retrieves the local graph context, and selects the next action from edge options. This allows a small model to focus on local recognition and decision making instead of reasoning over the entire app screenshot and task trajectory from scratch. The runtime agent still remains flexible: when the current screen cannot be matched to a node or the desired action is not covered by existing edges, it falls back to a free-action planner. Figure~\ref{fig:overview} displays the workflow of the runtime GUI agent in blue blocks.

\subsection{Runtime Graph Retrieval}

Given a user task $\tau$ and the current screenshot $x_t$, the runtime agent first locates the current UI state in the app knowledge graph $G_{\mathcal{A}}=(V,E)$ constructed by UI-KOBE. Unlike exploration, where new nodes can be created, runtime agent treats the graph as a fixed knowledge source. The goal is therefore to identify the most relevant existing node rather than expand the graph. For each graph node $v \in V$, the runtime gets access to the semantic description, state schema, outgoing edges, and cached visual embedding. Given the current screenshot, the agent computes a visual representation and retrieves a small set of candidate nodes with similar reference screenshots. These candidates are then provided to a model as a constrained selection problem, where each option contains the semantic description and retrieval score. The model either selects the best-matching node or rejects all candidates if none correspond to the current screen.

This two-stage identification process combines efficient visual retrieval with model-based semantic verification. Visual retrieval narrows the search space, while the final selection step reduces errors caused by visually similar but functionally different screens. If no node is accepted, the agent marks the current step as graph-unmatched and invokes fallback planning.

\subsection{Graph-Guided Decision Making}

Once the current node $v_t$ is identified, the agent constructs a local action option list from the graph. The list consists of four types of options: task completion, self-loop actions, neighboring transitions, and free actions. Self-loop actions correspond to edges that modify the internal state of the current screen while preserving the same UI template. Neighboring transitions correspond to edges that move the app from the current node to another node. The free-action option allows the model to propose an action not covered by the graph. And the task completion option allows the agent to terminate the execution.

Formally, the runtime agent selects an option conditioned on the user task $\tau$, current screenshot $x_t$, identified node $v_t$, local outgoing edges $\mathcal{E}(v_t)$, and runtime memory $m_t$:
\[
o_t = \pi_{\theta}\left(\tau, x_t, v_t, \mathcal{E}(v_t), m_t\right),
\]
where $o_t$ denotes either a graph-supported option or a fallback free action. The local edge set $\mathcal{E}(v_t)$ contains self-loop edges and one-hop transitions from $v_t$. Each edge provides its instruction, target observation, and optional schema delta, informing the model what actions are available and what effects they are expected to produce.

After selecting an option, the agent sends its instruction to an action grounding model, which converts the current screenshot and instruction into an executable device action, such as tapping, typing, swiping, or pressing a system button. This separates high-level option selection from low-level action grounding, keeping each runtime decision narrow and interpretable.

\subsection{Runtime Memory and Task Progress}

The runtime agent maintains a lightweight memory module to track task progress across steps. The memory records completed instructions, extracted task-relevant information, and recent observations. For example, when the task requires finding a specific item, the memory may store whether a query has already been entered, whether a relevant result has appeared, or whether a confirmation message has been observed. This prevents the agent from repeatedly executing the same graph edge and helps it determine when the task has been completed. At each step, the agent performs a record stage before decision making. Given the current screenshot, task, and previous actions, the model extracts concise factual information relevant to the task. The extracted facts are added to memory and then used together with the local graph options during decision making. 

\subsection{Fallback Planning}

Graph guidance may be unavailable when the current screen is not covered by the graph, when node retrieval is uncertain, or when the graph does not contain the action needed for the current task. In these cases, the agent does not directly send the entire user task to the action grounding model. Instead, it invokes a fallback planner that produces a concrete one-step instruction based on the current screenshot, task, action history, and memory as an ordinary GUI agent.

The fallback planner preserves the same decision interface as graph-guided execution: it outputs only the next immediate instruction, which is then grounded into a device action by the action model. This prevents the action grounding model from being responsible for long-horizon planning and keeps execution robust even outside graph-supported states. When the app returns to a known screen in a later step, the agent resumes graph-guided decision making through the normal identify-and-decide loop.

% \subsection{Discussion}

% The graph-guided GUI agent converts app interaction into a sequence of locally grounded decisions. Compared with end-to-end planning from screenshots alone, this formulation provides three advantages for small models. First, the graph constrains the action space to interactions that have been observed during exploration, reducing decision ambiguity. Second, target observations and schema deltas provide explicit expectations about the effect of each action, improving task progress estimation. Third, the runtime trajectory becomes interpretable because each executed action can be traced back to a graph node and edge. As a result, the agent can better leverage lightweight models for GUI task execution while retaining flexibility through fallback planning.

\begin{table}[t]
\centering
\renewcommand{\arraystretch}{1.05}
\setlength{\tabcolsep}{8pt}
\resizebox{0.98\linewidth}{!}{%
\begin{tabular}{L{4cm} | C{4cm}}
\toprule
\textbf{Statistic} & \textbf{Average per App} \\
\midrule
Nodes & 54 \\
Edges & 226 \\
Construction Cost & \$ 6.2 \\
Construction Time & 3.2 hours \\
\bottomrule
\end{tabular}
}
\caption{Statistics of UI-KOBE knowledge graph construction. We report the average number of audited nodes and edges per app, together with the average offline construction cost and time. Each graph is built once per app and reused across runtime tasks.}
\label{tab:graph_stats}
\end{table}

\section{Experiments}

\begin{table*}[t]
\centering
\renewcommand{\arraystretch}{1.05}
\setlength{\tabcolsep}{4pt}
\resizebox{0.98\textwidth}{!}{%
    \begin{tabular}{C{3.5cm}| L{6.5cm} | C{4.2cm} | C{2.7cm}}
        \toprule
        \textbf{Agent Type} & \textbf{Agent} & \textbf{Size / Model} & \textbf{Success Rate}  \\
        \midrule
        \multirow{18}{*}{\textbf{Single Model}} 
        & ScaleCUA-3B~\cite{liu2025scalecuascalingopensourcecomputer} & 3B & 23.7 \\
        & Ferret-UI-Lite-3B~\cite{yang2025ferretuilitelessonsbuilding} & 3B & 28.0 \\
        & UI-Tars-1.5-7B~\cite{qin2025uitarspioneeringautomatedgui} & 7B & 30.0 \\
        & UI-Tars-7B~\cite{qin2025uitarspioneeringautomatedgui} & 7B & 33.0 \\
        & Qwen3-VL-2B~\cite{bai2025qwen3vltechnicalreport} & 2B & 36.4 \\
        & UI-Tars-72B~\cite{qin2025uitarspioneeringautomatedgui} & 72B & 46.6 \\
        & Qwen3-VL-8B~\cite{bai2025qwen3vltechnicalreport} & 8B & 47.6 \\
        & MAI-UI-2B~\cite{zhou2025maiuitechnicalreportrealworld} & 2B & 49.1 \\
        & UI-Venus-7B~\cite{gu2025uivenustechnicalreportbuilding} & 7B & 49.1 \\
        & Qwen3-VL-32B~\cite{bai2025qwen3vltechnicalreport} & 32B & 57.3 \\
        & Qwen3.5-9B~\cite{qwen3.5} & 9B & 57.8 \\
        & Qwen3.5-4B~\cite{qwen3.5} & 4B & 58.6 \\
        & Qwen3-VL-235B-A22B~\cite{bai2025qwen3vltechnicalreport} & 235B & 63.7 \\
        & UI-Venus-72B~\cite{venusteam2026uivenus15technicalreport} & 72B & 65.9 \\
        & GUI-Owl-7B~\cite{ye2025mobileagentv3fundamentalagentsgui} & 7B & 66.4 \\
        & Qwen3.5-plus~\cite{qwen3.5} & 397B & 66.8 \\
        & Step-GUI-8B~\cite{yan2025stepguitechnicalreport} & 8B & 67.7 \\
        & MAI-UI-8B~\cite{zhou2025maiuitechnicalreportrealworld} & 8B & 70.7 \\
        & MAI-UI-32B~\cite{zhou2025maiuitechnicalreportrealworld} & 32B & 73.3 \\
        & MAI-UI-235B-A22B~\cite{zhou2025maiuitechnicalreportrealworld} & 235B & 76.7 \\
        \midrule
        \multirow{5}{*}{\textbf{Agentic Framework}} 
        & GUI-Explorer~\cite{sun2025guixplore} & GPT-4o & 47.4 \\
        & Agent-S2~\cite{agashe2025agents2compositionalgeneralistspecialist} & Claude-3.7-sonnet & 54.3 \\
        & V-Droid~\cite{dai2026vdroid} & V-Droid-8B & 59.5 \\
        & MobileUse~\cite{li2025mobileuseguiagenthierarchical} & Qwen2.5-VL-72B & 62.9 \\
        & Mobile-Agent-v3~\cite{ye2025mobileagentv3fundamentalagentsgui} & GUI-Owl-32B & 73.3 \\
        \midrule
        \multirow{3}{*}{\textbf{Our}} 
        & \textbf{UI-KOBE} & Qwen3.5-4B & 70.7 \\
        & \textbf{UI-KOBE} & Qwen3.5-9B & 72.4 \\
        & \textbf{UI-KOBE} & Qwen3.5-Plus & 77.6 \\
        \bottomrule
    \end{tabular}%
}
\caption{Results on AndroidWorld. UI-KOBE consistently improves GUI task success across different runtime model scales, achieving competitive performance with a 4B model and the best overall success rate with Qwen3.5-Plus compared with representative single-model agents and agentic frameworks.}
\label{tab:android-world}
\end{table*}

\subsection{Experimental Settings}

\label{sec:exp_settings}

\paragraph{Benchmarks.}

We evaluate UI-KOBE on two mobile GUI benchmarks: \textbf{AndroidWorld}~\cite{rawles2025androidworlddynamicbenchmarkingenvironment} and \textbf{A3}~\cite{chai2026a3androidagentarena}. AndroidWorld evaluates interactive Android task automation, where agents execute step-by-step device actions from natural-language instructions. A3 further evaluates agents on realistic online mobile app tasks with dynamic UI states. For AndroidWorld, we report task success rate (SR). For A3, we report essential-state achievement rate (ESAR) and overall success rate (Overall SR), where ESAR measures fine-grained task progress and Overall SR measures full task completion.

\paragraph{Models.}

During UI-KOBE exploration, we use \texttt{Qwen3.5-Plus} for action grounding and \texttt{GPT-5.4} for page description, action planning, node verification, and graph auditing. We use \texttt{Gemini-Embedding-2} for graph retrieval and node matching. For runtime execution, we instantiate the graph-guided agent with \texttt{Qwen3.5-4B}, \texttt{Qwen3.5-9B}, and \texttt{Qwen3.5-Plus}, covering lightweight to stronger model scales.

\subsection{Graph Statistics}

\label{sec:graph_stats}

Before evaluating task performance, we summarize the app knowledge graphs constructed by UI-KOBE on AndroidWorld and A3. Each app is explored at 300 steps, and we report the average number of audited nodes and edges, as well as the average cost and time required for graph construction. As shown in Table~\ref{tab:graph_stats}, UI-KOBE constructs compact app-level graphs that can be reused across tasks. Although graph construction introduces a one-time overhead, this cost is amortized over repeated task executions within the same app. The resulting graph provides the runtime agent with explicit app structure and transition knowledge, enabling graph-guided decision making without repeated exploration during task execution.

\begin{table*}[t]
    \centering
    \renewcommand{\arraystretch}{1.05}
    \setlength{\tabcolsep}{4pt}
    \resizebox{0.98\linewidth}{!}{%
    \begin{tabular}{C{3.5cm} | L{5.5cm} | C{3cm} | C{2.7cm} | C{2.7cm}}
    \toprule
        \textbf{Agent Type} & \textbf{Agent} & \textbf{Size / Model} & \textbf{ESAR} & \textbf{Overall SR} \\
        \midrule
        \multirow{10}{*}{\textbf{Single Model}} 
        & Qwen2.5-VL~\cite{bai2025qwen25vltechnicalreport}      & 7B & 14.2 & 3 \\
        & UI-TARS-1.5~\cite{qin2025uitarspioneeringautomatedgui}     & 7B & 28.2 & 12 \\
        & UI-Genie~\cite{xiao2025uigenieselfimprovingapproachiteratively}        & 7B & 32.1 & 13 \\
        & GUI-OWL~\cite{ye2025mobileagentv3fundamentalagentsgui}         & 7B & 32.0 & 14 \\
        & Qwen3-VL~\cite{bai2025qwen3vltechnicalreport}        & 8B & 38.2 & 17 \\
        & UI-Venus~\cite{gu2025uivenustechnicalreportbuilding}        & 7B & 32.0 & 20 \\
        & Qwen3-VL~\cite{bai2025qwen3vltechnicalreport}        & 30B-A3B & 45.6 & 27 \\
        & Qwen3.5~\cite{qwen3.5}         & 4B & 43.7 & 26 \\
        & Qwen3.5~\cite{qwen3.5}          & 9B & 51.7 & 31 \\
        & Qwen3.5-plus~\cite{qwen3.5}     & 397B & 67.9 & 52 \\
        \midrule
        \multirow{3}{*}{\textbf{Agentic Framework}} 
        & Mobile-Use~\cite{li2025mobileuseguiagenthierarchical}      & Qwen2.5-VL-7B & 39.5 & 16 \\
        & T3A~\cite{rawles2025androidworlddynamicbenchmarkingenvironment}             & Qwen2.5-VL-7B & 30.7 & 15 \\
        & T3A~\cite{rawles2025androidworlddynamicbenchmarkingenvironment}             & Gemini-2.5-pro & 66.4 & 53 \\
        \midrule
        \multirow{3}{*}{\textbf{Our}} 
        & \textbf{UI-KOBE} & Qwen3.5-4B & 71.5 & 61 \\
        & \textbf{UI-KOBE} & Qwen3.5-9B & 75.7 & 67 \\
        & \textbf{UI-KOBE} & Qwen3.5-Plus & 84.8 & 78 \\
    \bottomrule
    \end{tabular}%
    }
    \caption{Results on A3. UI-KOBE substantially improves both essential state achievement rate (ESAR) and overall task success rate (Overall SR) across different runtime model scales, outperforming representative single-model agents and agentic frameworks.}
    \label{tab:a3}
\end{table*}

\subsection{Main Results and Analysis}

\label{sec:main_results}

\paragraph{Results on AndroidWorld.}

Table~\ref{tab:android-world} shows the results on AndroidWorld. UI-KOBE achieves strong performance across all three runtime models. With \texttt{Qwen3.5-4B}, UI-KOBE reaches a success rate of 70.7\%, substantially outperforming the same backbone model without graph guidance, which achieves 58.6\%. This demonstrates that reusable app graph knowledge can significantly improve lightweight GUI agents without increasing model size. With \texttt{Qwen3.5-9B}, UI-KOBE further improves to 72.4\%, and with \texttt{Qwen3.5-Plus}, it reaches 77.6\%, outperforming all compared single-model agents and agentic frameworks in the table. These results suggest that UI-KOBE is effective not only for small models but also for stronger models. Notably, UI-KOBE with \texttt{Qwen3.5-4B} achieves performance comparable to or better than many much larger single-model agents and agentic systems. For example, it outperforms \texttt{Qwen3.5-Plus} without graph guidance (66.8\%) and Mobile-Agent-v3 with GUI-Owl-32B (73.3\%) is only slightly higher than the 4B UI-KOBE setting while using a much larger base model. This indicates that graph guidance can compensate for limited model capacity by reducing the burden of end-to-end GUI planning.

\paragraph{Results on A3.}

Table~\ref{tab:a3} presents results on A3. UI-KOBE again provides consistent improvements across model scales. With \texttt{Qwen3.5-4B}, UI-KOBE achieves 71.5 ESAR and 61 Overall SR, compared with 43.7 ESAR and 26 Overall SR for the original \texttt{Qwen3.5-4B}. This large improvement shows that graph guidance is especially beneficial in realistic online app scenarios, where small models often struggle with long-horizon planning and dynamic UI states. The gains remain significant for larger models. UI-KOBE with \texttt{Qwen3.5-9B} achieves 75.7 ESAR and 67 Overall SR, improving over the original \texttt{Qwen3.5-9B} by 19.8 ESAR and 36 Overall SR. With \texttt{Qwen3.5-Plus}, UI-KOBE reaches 84.8 ESAR and 78 Overall SR, outperforming both the original \texttt{Qwen3.5-Plus} and the strongest agentic framework baseline, T3A with Gemini-2.5-pro. These results further verify that app-specific graph knowledge improves both fine-grained task progress and full task completion.

\paragraph{Effect of Graph Guidance.}

Across both benchmarks, UI-KOBE improves GUI task execution by replacing end-to-end planning with graph-guided step-by-step decision making. Without graph guidance, the runtime model must infer the current app state, possible navigation paths, and task progress directly from screenshots. In contrast, UI-KOBE provides an explicit app knowledge graph that indicates the likely current state, available transitions, and expected action effects. This reduces decision ambiguity and makes execution more reliable, especially for smaller models. The results also show that graph guidance is complementary to model scale: larger models still benefit, while the relative improvement is particularly strong for lightweight models. UI-KOBE does introduce an exploration cost, averaging \$6.2 and 6.4 hours per app as shown in Table~\ref{tab:graph_stats}. However, this cost is paid only once per app and can be amortized across future tasks in the same application. The performance gains in Tables~\ref{tab:android-world} and~\ref{tab:a3} suggest that this trade-off is practical, especially in repeated-use scenarios where reusable app knowledge can improve runtime execution without increasing model size or requiring task-specific training. A more detailed error analysis is provided in Appendix~\ref{app:error-study}.

\section{Conclusion}

We present \textbf{UI-KOBE}, an exploration method for constructing reusable app knowledge graphs. By autonomously exploring mobile apps, UI-KOBE captures UI states, executable transitions, and interaction knowledge that can later guide GUI agents during task execution. Based on this graph, our graph-guided runtime agent reduces the burden of end-to-end planning by turning task execution into step-by-step decisions supported by app-specific knowledge. Experiments on AndroidWorld and A3 show that UI-KOBE improves GUI task performance across different model scales, with particularly strong gains for lightweight models. These results suggest that reusable app knowledge is a promising direction for building efficient and deployable GUI agents.

\clearpage
\section*{Limitations}

UI-KOBE has several limitations that we plan to address in future work. First, the constructed graph is app-version dependent. When an application introduces major UI or navigation changes, the existing graph may become partially outdated and require incremental repair or re-exploration. Second, although our goal is to support lightweight on-device GUI agents, the current system still relies on an external embedding model for graph retrieval and node matching, which prevents a fully local deployment. Third, our experiments focus on mobile applications, leaving the effectiveness of UI-KOBE on websites and PC applications unverified. Extending graph construction, graph maintenance, and graph-guided execution to these broader GUI environments remains an important next step.

\bibliography{custom}

@inproceedings{chai-etal-2025-amex,
    title = "{AMEX}: Android Multi-annotation Expo Dataset for Mobile {GUI} Agents",
    author = "Chai, Yuxiang  and
      Huang, Siyuan  and
      Niu, Yazhe  and
      Xiao, Han  and
      Liu, Liang  and
      Wang, Guozhi  and
      Zhang, Dingyu  and
      Ren, Shuai  and
      Li, Hongsheng",
    editor = "Che, Wanxiang  and
      Nabende, Joyce  and
      Shutova, Ekaterina  and
      Pilehvar, Mohammad Taher",
    booktitle = "Findings of the Association for Computational Linguistics: ACL 2025",
    month = jul,
    year = "2025",
    address = "Vienna, Austria",
    publisher = "Association for Computational Linguistics",
    url = "https://aclanthology.org/2025.findings-acl.110/",
    doi = "10.18653/v1/2025.findings-acl.110",
    pages = "2138--2156",
    ISBN = "979-8-89176-256-5",
}

@misc{rawles2025androidworlddynamicbenchmarkingenvironment,
      title={AndroidWorld: A Dynamic Benchmarking Environment for Autonomous Agents}, 
      author={Christopher Rawles and Sarah Clinckemaillie and Yifan Chang and Jonathan Waltz and Gabrielle Lau and Marybeth Fair and Alice Li and William Bishop and Wei Li and Folawiyo Campbell-Ajala and Daniel Toyama and Robert Berry and Divya Tyamagundlu and Timothy Lillicrap and Oriana Riva},
      year={2025},
      eprint={2405.14573},
      archivePrefix={arXiv},
      primaryClass={cs.AI},
      url={https://arxiv.org/abs/2405.14573}, 
}

@misc{qin2025uitarspioneeringautomatedgui,
      title={UI-TARS: Pioneering Automated GUI Interaction with Native Agents}, 
      author={Yujia Qin and Yining Ye and Junjie Fang and Haoming Wang and Shihao Liang and Shizuo Tian and Junda Zhang and Jiahao Li and Yunxin Li and Shijue Huang and Wanjun Zhong and Kuanye Li and Jiale Yang and Yu Miao and Woyu Lin and Longxiang Liu and Xu Jiang and Qianli Ma and Jingyu Li and Xiaojun Xiao and Kai Cai and Chuang Li and Yaowei Zheng and Chaolin Jin and Chen Li and Xiao Zhou and Minchao Wang and Haoli Chen and Zhaojian Li and Haihua Yang and Haifeng Liu and Feng Lin and Tao Peng and Xin Liu and Guang Shi},
      year={2025},
      eprint={2501.12326},
      archivePrefix={arXiv},
      primaryClass={cs.AI},
      url={https://arxiv.org/abs/2501.12326}, 
}

@misc{liu2025llmpoweredguiagentsphone,
      title={LLM-Powered GUI Agents in Phone Automation: Surveying Progress and Prospects}, 
      author={Guangyi Liu and Pengxiang Zhao and Yaozhen Liang and Liang Liu and Yaxuan Guo and Han Xiao and Weifeng Lin and Yuxiang Chai and Yue Han and Shuai Ren and Hao Wang and Xiaoyu Liang and WenHao Wang and Tianze Wu and Zhengxi Lu and Siheng Chen and LiLinghao and Hao Wang and Guanjing Xiong and Yong Liu and Hongsheng Li},
      year={2025},
      eprint={2504.19838},
      archivePrefix={arXiv},
      primaryClass={cs.HC},
      url={https://arxiv.org/abs/2504.19838}, 
}

@misc{bai2025qwen3vltechnicalreport,
      title={Qwen3-VL Technical Report}, 
      author={Shuai Bai and Yuxuan Cai and Ruizhe Chen and Keqin Chen and Xionghui Chen and Zesen Cheng and Lianghao Deng and Wei Ding and Chang Gao and Chunjiang Ge and Wenbin Ge and Zhifang Guo and Qidong Huang and Jie Huang and Fei Huang and Binyuan Hui and Shutong Jiang and Zhaohai Li and Mingsheng Li and Mei Li and Kaixin Li and Zicheng Lin and Junyang Lin and Xuejing Liu and Jiawei Liu and Chenglong Liu and Yang Liu and Dayiheng Liu and Shixuan Liu and Dunjie Lu and Ruilin Luo and Chenxu Lv and Rui Men and Lingchen Meng and Xuancheng Ren and Xingzhang Ren and Sibo Song and Yuchong Sun and Jun Tang and Jianhong Tu and Jianqiang Wan and Peng Wang and Pengfei Wang and Qiuyue Wang and Yuxuan Wang and Tianbao Xie and Yiheng Xu and Haiyang Xu and Jin Xu and Zhibo Yang and Mingkun Yang and Jianxin Yang and An Yang and Bowen Yu and Fei Zhang and Hang Zhang and Xi Zhang and Bo Zheng and Humen Zhong and Jingren Zhou and Fan Zhou and Jing Zhou and Yuanzhi Zhu and Ke Zhu},
      year={2025},
      eprint={2511.21631},
      archivePrefix={arXiv},
      primaryClass={cs.CV},
      url={https://arxiv.org/abs/2511.21631}, 
}

@misc{wang2025guiagentsfoundationmodels,
      title={GUI Agents with Foundation Models: A Comprehensive Survey}, 
      author={Shuai Wang and Weiwen Liu and Jingxuan Chen and Yuqi Zhou and Weinan Gan and Xingshan Zeng and Yuhan Che and Shuai Yu and Xinlong Hao and Kun Shao and Bin Wang and Chuhan Wu and Yasheng Wang and Ruiming Tang and Jianye Hao},
      year={2025},
      eprint={2411.04890},
      archivePrefix={arXiv},
      primaryClass={cs.AI},
      url={https://arxiv.org/abs/2411.04890}, 
}

@misc{ye2025mobileagentv3fundamentalagentsgui,
      title={Mobile-Agent-v3: Fundamental Agents for GUI Automation}, 
      author={Jiabo Ye and Xi Zhang and Haiyang Xu and Haowei Liu and Junyang Wang and Zhaoqing Zhu and Ziwei Zheng and Feiyu Gao and Junjie Cao and Zhengxi Lu and Jitong Liao and Qi Zheng and Fei Huang and Jingren Zhou and Ming Yan},
      year={2025},
      eprint={2508.15144},
      archivePrefix={arXiv},
      primaryClass={cs.AI},
      url={https://arxiv.org/abs/2508.15144}, 
}

@inproceedings{hu-etal-2025-os,
    title = "{OS} Agents: A Survey on {MLLM}-based Agents for Computer, Phone and Browser Use",
    author = "Hu, Xueyu  and
      Xiong, Tao  and
      Yi, Biao  and
      Wei, Zishu  and
      Xiao, Ruixuan  and
      Chen, Yurun  and
      Ye, Jiasheng  and
      Tao, Meiling  and
      Zhou, Xiangxin  and
      Zhao, Ziyu  and
      Li, Yuhuai  and
      Xu, Shengze  and
      Wang, Shenzhi  and
      Xu, Xinchen  and
      Qiao, Shuofei  and
      Wang, Zhaokai  and
      Kuang, Kun  and
      Zeng, Tieyong  and
      Wang, Liang  and
      Li, Jiwei  and
      Jiang, Yuchen Eleanor  and
      Zhou, Wangchunshu  and
      Wang, Guoyin  and
      Yin, Keting  and
      Zhao, Zhou  and
      Yang, Hongxia  and
      Wu, Fan  and
      Zhang, Shengyu  and
      Wu, Fei",
    editor = "Che, Wanxiang  and
      Nabende, Joyce  and
      Shutova, Ekaterina  and
      Pilehvar, Mohammad Taher",
    booktitle = "Proceedings of the 63rd Annual Meeting of the Association for Computational Linguistics (Volume 1: Long Papers)",
    month = jul,
    year = "2025",
    address = "Vienna, Austria",
    publisher = "Association for Computational Linguistics",
    url = "https://aclanthology.org/2025.acl-long.369/",
    doi = "10.18653/v1/2025.acl-long.369",
    pages = "7436--7465",
    ISBN = "979-8-89176-251-0",
}

@misc{bai2025qwen25vltechnicalreport,
      title={Qwen2.5-VL Technical Report}, 
      author={Shuai Bai and Keqin Chen and Xuejing Liu and Jialin Wang and Wenbin Ge and Sibo Song and Kai Dang and Peng Wang and Shijie Wang and Jun Tang and Humen Zhong and Yuanzhi Zhu and Mingkun Yang and Zhaohai Li and Jianqiang Wan and Pengfei Wang and Wei Ding and Zheren Fu and Yiheng Xu and Jiabo Ye and Xi Zhang and Tianbao Xie and Zesen Cheng and Hang Zhang and Zhibo Yang and Haiyang Xu and Junyang Lin},
      year={2025},
      eprint={2502.13923},
      archivePrefix={arXiv},
      primaryClass={cs.CV},
      url={https://arxiv.org/abs/2502.13923}, 
}

@misc{zhang2023appagentmultimodalagentssmartphone,
      title={AppAgent: Multimodal Agents as Smartphone Users}, 
      author={Chi Zhang and Zhao Yang and Jiaxuan Liu and Yucheng Han and Xin Chen and Zebiao Huang and Bin Fu and Gang Yu},
      year={2023},
      eprint={2312.13771},
      archivePrefix={arXiv},
      primaryClass={cs.CV},
      url={https://arxiv.org/abs/2312.13771}, 
}

@misc{xiao2025uigenieselfimprovingapproachiteratively,
      title={UI-Genie: A Self-Improving Approach for Iteratively Boosting MLLM-based Mobile GUI Agents}, 
      author={Han Xiao and Guozhi Wang and Yuxiang Chai and Zimu Lu and Weifeng Lin and Hao He and Lue Fan and Liuyang Bian and Rui Hu and Liang Liu and Shuai Ren and Yafei Wen and Xiaoxin Chen and Aojun Zhou and Hongsheng Li},
      year={2025},
      eprint={2505.21496},
      archivePrefix={arXiv},
      primaryClass={cs.CL},
      url={https://arxiv.org/abs/2505.21496}, 
}

@misc{gu2025uivenustechnicalreportbuilding,
      title={UI-Venus Technical Report: Building High-performance UI Agents with RFT}, 
      author={Zhangxuan Gu and Zhengwen Zeng and Zhenyu Xu and Xingran Zhou and Shuheng Shen and Yunfei Liu and Beitong Zhou and Changhua Meng and Tianyu Xia and Weizhi Chen and Yue Wen and Jingya Dou and Fei Tang and Jinzhen Lin and Yulin Liu and Zhenlin Guo and Yichen Gong and Heng Jia and Changlong Gao and Yuan Guo and Yong Deng and Zhenyu Guo and Liang Chen and Weiqiang Wang},
      year={2025},
      eprint={2508.10833},
      archivePrefix={arXiv},
      primaryClass={cs.CV},
      url={https://arxiv.org/abs/2508.10833}, 
}

@misc{liu2025infiguir1advancingmultimodalgui,
      title={InfiGUI-R1: Advancing Multimodal GUI Agents from Reactive Actors to Deliberative Reasoners}, 
      author={Yuhang Liu and Pengxiang Li and Congkai Xie and Xavier Hu and Xiaotian Han and Shengyu Zhang and Hongxia Yang and Fei Wu},
      year={2025},
      eprint={2504.14239},
      archivePrefix={arXiv},
      primaryClass={cs.AI},
      url={https://arxiv.org/abs/2504.14239}, 
}

@article{xiao2026ui,
  title={UI-Mem: Self-Evolving Experience Memory for Online Reinforcement Learning in Mobile GUI Agents},
  author={Xiao, Han and Wang, Guozhi and Wang, Hao and Liu, Shilong and Chai, Yuxiang and Pan, Yue and Zhou, Yufeng and Chen, Xiaoxin and Wen, Yafei and Li, Hongsheng},
  journal={arXiv preprint arXiv:2602.05832},
  year={2026}
}

@article{yu2026graphpilot,
  title={GraphPilot: GUI Task Automation with One-Step LLM Reasoning Powered by Knowledge Graph},
  author={Yu, Mingxian and Luo, Siqi and Chen, Xu},
  journal={arXiv preprint arXiv:2601.17418},
  year={2026}
}

@misc{guan2025kgragenhancingguiagent,
      title={KG-RAG: Enhancing GUI Agent Decision-Making via Knowledge Graph-Driven Retrieval-Augmented Generation}, 
      author={Ziyi Guan and Jason Chun Lok Li and Zhijian Hou and Pingping Zhang and Donglai Xu and Yuzhi Zhao and Mengyang Wu and Jinpeng Chen and Thanh-Toan Nguyen and Pengfei Xian and Wenao Ma and Shengchao Qin and Graziano Chesi and Ngai Wong},
      year={2025},
      eprint={2509.00366},
      archivePrefix={arXiv},
      primaryClass={cs.MA},
      url={https://arxiv.org/abs/2509.00366}, 
}

@inproceedings{sun2025guixplore,
  title={Gui-xplore: Empowering generalizable gui agents with one exploration},
  author={Sun, Yuchen and Zhao, Shanhui and Yu, Tao and Wen, Hao and Va, Samith and Xu, Mengwei and Li, Yuanchun and Zhang, Chongyang},
  booktitle={Proceedings of the computer vision and pattern recognition conference},
  pages={19477--19486},
  year={2025}
}

@misc{zhou2025maiuitechnicalreportrealworld,
      title={MAI-UI Technical Report: Real-World Centric Foundation GUI Agents}, 
      author={Hanzhang Zhou and Xu Zhang and Panrong Tong and Jianan Zhang and Liangyu Chen and Quyu Kong and Chenglin Cai and Chen Liu and Yue Wang and Jingren Zhou and Steven Hoi},
      year={2025},
      eprint={2512.22047},
      archivePrefix={arXiv},
      primaryClass={cs.CV},
      url={https://arxiv.org/abs/2512.22047}, 
}

@misc{venusteam2026uivenus15technicalreport,
      title={UI-Venus-1.5 Technical Report}, 
      author={Venus Team and Changlong Gao and Zhangxuan Gu and Yulin Liu and Xinyu Qiu and Shuheng Shen and Yue Wen and Tianyu Xia and Zhenyu Xu and Zhengwen Zeng and Beitong Zhou and Xingran Zhou and Weizhi Chen and Sunhao Dai and Jingya Dou and Yichen Gong and Yuan Guo and Zhenlin Guo and Feng Li and Qian Li and Jinzhen Lin and Yuqi Zhou and Linchao Zhu and Liang Chen and Zhenyu Guo and Changhua Meng and Weiqiang Wang},
      year={2026},
      eprint={2602.09082},
      archivePrefix={arXiv},
      primaryClass={cs.CV},
      url={https://arxiv.org/abs/2602.09082}, 
}

@misc{xu2026mobileagentv35multiplatformfundamentalgui,
      title={Mobile-Agent-v3.5: Multi-platform Fundamental GUI Agents}, 
      author={Haiyang Xu and Xi Zhang and Haowei Liu and Junyang Wang and Zhaozai Zhu and Shengjie Zhou and Xuhao Hu and Feiyu Gao and Junjie Cao and Zihua Wang and Zhiyuan Chen and Jitong Liao and Qi Zheng and Jiahui Zeng and Ze Xu and Shuai Bai and Junyang Lin and Jingren Zhou and Ming Yan},
      year={2026},
      eprint={2602.16855},
      archivePrefix={arXiv},
      primaryClass={cs.AI},
      url={https://arxiv.org/abs/2602.16855}, 
}

@misc{lu2025uir1enhancingefficientaction,
      title={UI-R1: Enhancing Efficient Action Prediction of GUI Agents by Reinforcement Learning}, 
      author={Zhengxi Lu and Yuxiang Chai and Yaxuan Guo and Xi Yin and Liang Liu and Hao Wang and Han Xiao and Shuai Ren and Guanjing Xiong and Hongsheng Li},
      year={2025},
      eprint={2503.21620},
      archivePrefix={arXiv},
      primaryClass={cs.AI},
      url={https://arxiv.org/abs/2503.21620}, 
}

@misc{yang2025ferretuilitelessonsbuilding,
      title={Ferret-UI Lite: Lessons from Building Small On-Device GUI Agents}, 
      author={Zhen Yang and Zi-Yi Dou and Di Feng and Forrest Huang and Anh Nguyen and Keen You and Omar Attia and Yuhao Yang and Michael Feng and Haotian Zhang and Ram Ramrakhya and Chao Jia and Jeffrey Nichols and Alexander Toshev and Yinfei Yang and Zhe Gan},
      year={2025},
      eprint={2509.26539},
      archivePrefix={arXiv},
      primaryClass={cs.CV},
      url={https://arxiv.org/abs/2509.26539}, 
}

@misc{wen2024autodroidllmpoweredtaskautomation,
      title={AutoDroid: LLM-powered Task Automation in Android}, 
      author={Hao Wen and Yuanchun Li and Guohong Liu and Shanhui Zhao and Tao Yu and Toby Jia-Jun Li and Shiqi Jiang and Yunhao Liu and Yaqin Zhang and Yunxin Liu},
      year={2024},
      eprint={2308.15272},
      archivePrefix={arXiv},
      primaryClass={cs.AI},
      url={https://arxiv.org/abs/2308.15272}, 
}

@misc{chai2026a3androidagentarena,
      title={A3: Android Agent Arena for Mobile GUI Agents with Essential-State Procedural Evaluation}, 
      author={Yuxiang Chai and Shunye Tang and Han Xiao and Weifeng Lin and Hanhao Li and Jiayu Zhang and Liang Liu and Pengxiang Zhao and Guangyi Liu and Guozhi Wang and Shuai Ren and Rongduo Han and Haining Zhang and Siyuan Huang and Hongsheng Li},
      year={2026},
      eprint={2501.01149},
      archivePrefix={arXiv},
      primaryClass={cs.AI},
      url={https://arxiv.org/abs/2501.01149}, 
}

@misc{qwen3.5,
    title  = {{Qwen3.5}: Towards Native Multimodal Agents},
    author = {{Qwen Team}},
    month  = {February},
    year   = {2026},
    url    = {https://qwen.ai/blog?id=qwen3.5}
}

@misc{li2025mobileuseguiagenthierarchical,
      title={MobileUse: A GUI Agent with Hierarchical Reflection for Autonomous Mobile Operation}, 
      author={Ning Li and Xiangmou Qu and Jiamu Zhou and Jun Wang and Muning Wen and Kounianhua Du and Xingyu Lou and Qiuying Peng and Jun Wang and Weinan Zhang},
      year={2025},
      eprint={2507.16853},
      archivePrefix={arXiv},
      primaryClass={cs.RO},
      url={https://arxiv.org/abs/2507.16853}, 
}

@misc{liu2025scalecuascalingopensourcecomputer,
      title={ScaleCUA: Scaling Open-Source Computer Use Agents with Cross-Platform Data}, 
      author={Zhaoyang Liu and Jingjing Xie and Zichen Ding and Zehao Li and Bowen Yang and Zhenyu Wu and Xuehui Wang and Qiushi Sun and Shi Liu and Weiyun Wang and Shenglong Ye and Qingyun Li and Xuan Dong and Yue Yu and Chenyu Lu and YunXiang Mo and Yao Yan and Zeyue Tian and Xiao Zhang and Yuan Huang and Yiqian Liu and Weijie Su and Gen Luo and Xiangyu Yue and Biqing Qi and Kai Chen and Bowen Zhou and Yu Qiao and Qifeng Chen and Wenhai Wang},
      year={2025},
      eprint={2509.15221},
      archivePrefix={arXiv},
      primaryClass={cs.CV},
      url={https://arxiv.org/abs/2509.15221}, 
}

@misc{yan2025stepguitechnicalreport,
      title={Step-GUI Technical Report}, 
      author={Haolong Yan and Jia Wang and Xin Huang and Yeqing Shen and Ziyang Meng and Zhimin Fan and Kaijun Tan and Jin Gao and Lieyu Shi and Mi Yang and Shiliang Yang and Zhirui Wang and Brian Li and Kang An and Chenyang Li and Lei Lei and Mengmeng Duan and Danxun Liang and Guodong Liu and Hang Cheng and Hao Wu and Jie Dong and Junhao Huang and Mei Chen and Renjie Yu and Shunshan Li and Xu Zhou and Yiting Dai and Yineng Deng and Yingdan Liang and Zelin Chen and Wen Sun and Chengxu Yan and Chunqin Xu and Dong Li and Fengqiong Xiao and Guanghao Fan and Guopeng Li and Guozhen Peng and Hongbing Li and Hang Li and Hongming Chen and Jingjing Xie and Jianyong Li and Jingyang Zhang and Jiaju Ren and Jiayu Yuan and Jianpeng Yin and Kai Cao and Liang Zhao and Liguo Tan and Liying Shi and Mengqiang Ren and Min Xu and Manjiao Liu and Mao Luo and Mingxin Wan and Na Wang and Nan Wu and Ning Wang and Peiyao Ma and Qingzhou Zhang and Qiao Wang and Qinlin Zeng and Qiong Gao and Qiongyao Li and Shangwu Zhong and Shuli Gao and Shaofan Liu and Shisi Gao and Shuang Luo and Xingbin Liu and Xiaojia Liu and Xiaojie Hou and Xin Liu and Xuanti Feng and Xuedan Cai and Xuan Wen and Xianwei Zhu and Xin Liang and Xin Liu and Xin Zhou and Yifan Sui and Yingxiu Zhao and Yukang Shi and Yunfang Xu and Yuqing Zeng and Yixun Zhang and Zejia Weng and Zhonghao Yan and Zhiguo Huang and Zhuoyu Wang and Zihan Yan and Zheng Ge and Jing Li and Yibo Zhu and Binxing Jiao and Xiangyu Zhang and Daxin Jiang},
      year={2025},
      eprint={2512.15431},
      archivePrefix={arXiv},
      primaryClass={cs.CV},
      url={https://arxiv.org/abs/2512.15431}, 
}

@misc{agashe2025agents2compositionalgeneralistspecialist,
      title={Agent S2: A Compositional Generalist-Specialist Framework for Computer Use Agents}, 
      author={Saaket Agashe and Kyle Wong and Vincent Tu and Jiachen Yang and Ang Li and Xin Eric Wang},
      year={2025},
      eprint={2504.00906},
      archivePrefix={arXiv},
      primaryClass={cs.AI},
      url={https://arxiv.org/abs/2504.00906}, 
}

@misc{dai2026vdroid,
      title={Advancing Mobile GUI Agents: A Verifier-Driven Approach to Practical Deployment}, 
      author={Gaole Dai and Shiqi Jiang and Ting Cao and Yuanchun Li and Yuqing Yang and Rui Tan and Mo Li and Lili Qiu},
      year={2026},
      eprint={2503.15937},
      archivePrefix={arXiv},
      primaryClass={cs.AI},
      url={https://arxiv.org/abs/2503.15937}, 
}

\clearpage

\appendix

\section{Appendix}

\subsection{Empirical Study of Design Choices}
\label{app:empirical-study}

We further study several alternative design choices for graph retrieval and graph construction. These preliminary experiments help explain why UI-KOBE adopts screenshot-based node matching and one-step edge construction.

\paragraph{Text-based Node Identification.}
One alternative is to identify the current graph node using only page descriptions and text embeddings. In this setting, the runtime model first generates a textual description of the current screenshot, and the system retrieves the closest graph node by comparing this description with stored node descriptions. However, we find this strategy to be unstable when the graph construction model and runtime model differ. For example, descriptions generated by GPT models and Qwen models may follow different styles, levels of detail, and semantic emphasis, even for the same screen. As a result, text embeddings may retrieve an incorrect node despite the underlying UI state being visually identical. This motivates our use of screenshot-based embeddings with model-based verification for runtime node identification, which reduces sensitivity to description text distribution shifts.

\paragraph{Compound-action Edges.}
We also explored constructing edges from compound actions rather than single-step actions. A compound action corresponds to a multi-step instruction such as ``search for coffee shops,'' which may require tapping the search box, entering text, and pressing the search button. In this design, the graph stores the entire interaction as one high-level edge instead of recording separate edges such as ``tap the search box,'' ``type the query,'' and ``press search.'' While compound edges make the graph more compact, they introduce two issues. First, intermediate UI states and observations are skipped, causing useful information to be missing from the graph. Second, the action grounding model may fail to faithfully execute a compound instruction, especially when it requires multiple precise low-level steps. This can produce incorrect or incomplete transitions during exploration.

\subsection{Error Study and Additional Analysis}
\label{app:error-study}

Although UI-KOBE substantially improves GUI task execution, its performance is still below the ideal success rate. We analyze failed trajectories and identify two major sources of errors: graph construction errors and incomplete graph coverage.

\begin{figure*}[t]
    \centering
    \includegraphics[width=0.95\linewidth]{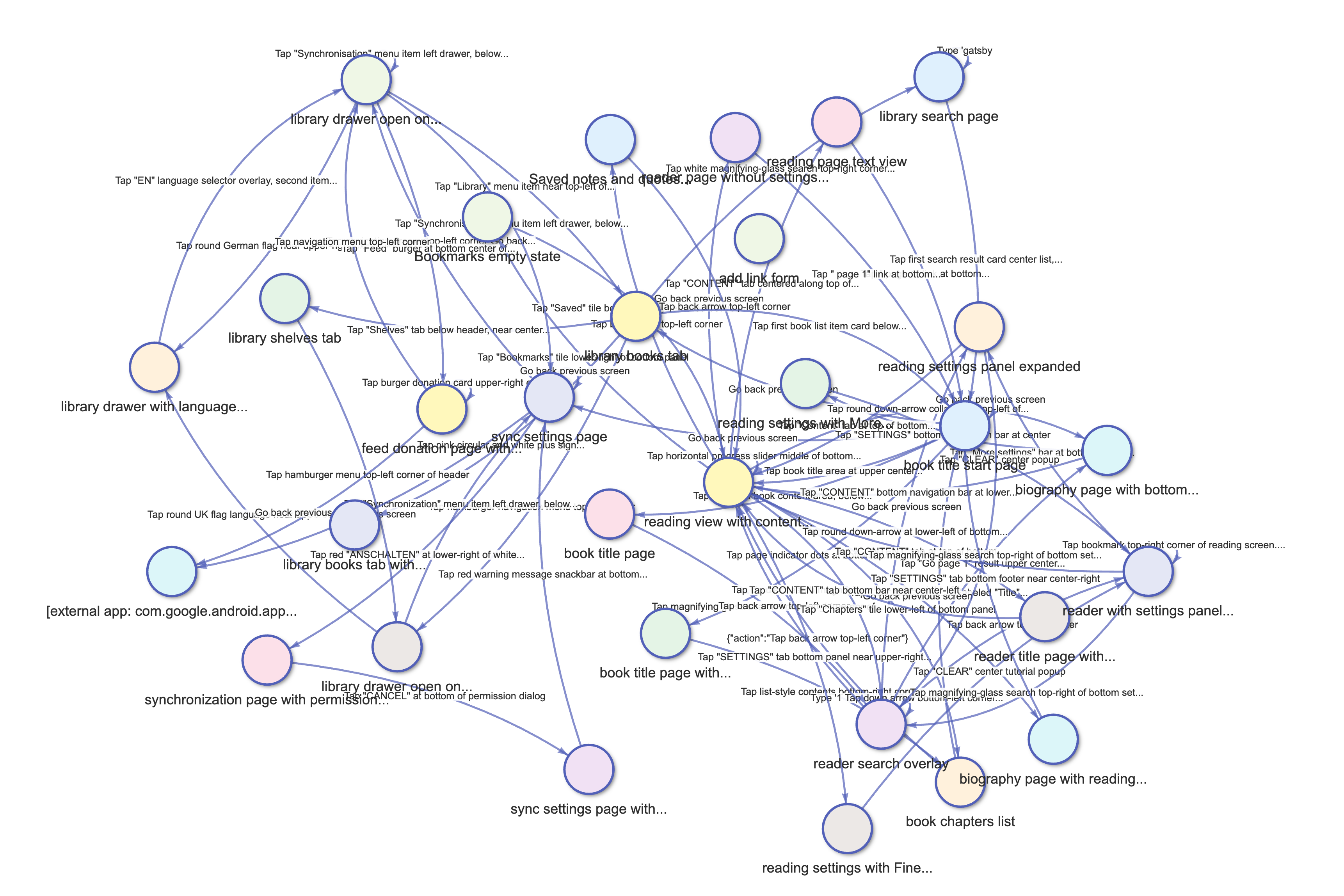}
    \caption{Visualization of the app knowledge graph constructed by UI-KOBE for the \textit{eboox} app. Nodes denote semantic UI states and directed edges denote executable transitions observed during exploration.}
    \label{fig:demo}
\end{figure*}

\paragraph{Graph Construction Errors.}
The first type of error comes from imperfections in the constructed app knowledge graph. In some cases, an edge may record an incorrect transition because the action grounding model does not fully follow the planned natural-language instruction during exploration. For example, the planner may intend to tap a specific UI element, but the grounded action may click a nearby or semantically different element. The resulting transition is then stored as if it were caused by the planned instruction, producing a misleading edge. Such errors are difficult to completely remove through post-hoc auditing, because the recorded source node, target node, and target observation may still appear locally plausible.

We also observe remaining duplicate nodes after graph auditing. UI-KOBE audits candidate nodes using semantic descriptions, reference screenshots, and outgoing actions, which can merge many duplicated states. However, some duplicate nodes remain when their descriptions are overly detailed. For instance, two visits to the same screen template may include different dynamic contents in their page descriptions, causing the audit model to treat them as different UI states. These unmerged duplicates can fragment outgoing transitions across multiple nodes, reducing the completeness of local graph context during runtime execution.

\paragraph{Incomplete Graph Coverage.}
The second major error source is incomplete exploration. Since UI-KOBE builds the graph through autonomous interaction at a limited step number for time and cost efficiency, some useful edges may not be discovered during exploration. When a runtime task requires an unexplored action, the agent cannot select it from the graph-supported action list and must instead rely on the fallback free-action planner. This explains why larger runtime models still achieve better performance than smaller ones under UI-KOBE: although all models benefit from graph guidance, stronger models are more capable when execution leaves the covered graph region.

A representative failure occurs when a task requires an action absent from the current node's outgoing edges. In this case, the \texttt{Qwen3.5-4B} agent produces an incorrect fallback instruction, leading the environment to a wrong node, which further compounds the error. In contrast, both \texttt{Qwen3.5-9B} and \texttt{Qwen3.5-Plus} produce the correct fallback instruction in the same case and successfully recover to graph-guided execution. This suggests that graph guidance reduces the burden of planning in covered states, but fallback planning remains an important bottleneck when graph coverage is incomplete.

\subsection{Qualitative Graph Visualization}

\label{app:graph-visualization}

To qualitatively illustrate the app knowledge graph constructed by UI-KOBE, we visualize the graph of the \textit{eboox} app in Figure~\ref{fig:demo}. We choose this app for visualization because its audited graph contains only around 30 nodes, making the full graph more readable; many other evaluated apps contain substantially larger graphs that are difficult to display clearly. Each node represents a semantic UI state, and each directed edge represents an executable transition discovered during exploration. The visualization shows that UI-KOBE captures both high-level navigation structures, such as moving between the library, drawer, reading view, search page, and settings pages, and local transitions around frequently used screens. This demonstrates how the constructed graph provides an explicit and interpretable representation of app behavior for downstream graph-guided execution.

\subsection{Potential Risks}

Like other GUI automation systems, UI-KOBE may trigger unintended actions if deployed without proper safeguards, especially for sensitive operations such as payments, messaging, account changes, or data deletion. Although the constructed graph captures general app behavior rather than user-specific data, exploration and execution should avoid private accounts or sensitive screens when possible. Practical deployments should include user confirmation for high-impact actions, sandboxed exploration, access control, and execution logs for auditing.

\subsection{AI Usage}

We used AI tools to assist paper writing, including language polishing and organization of technical descriptions. AI models are also part of the proposed system. 

\subsection{Licenses}

All models, datasets, and benchmarks used in this work are accessed and used in accordance with their respective licenses, terms of use, and intended research purposes. 

\end{document}